\documentclass{article}

    \usepackage[nonatbib, preprint]{neurips_2019}

\usepackage[utf8]{inputenc} 
\usepackage[T1]{fontenc}    
\usepackage{hyperref}       
\usepackage{url}            
\usepackage{booktabs}       
\usepackage{amsfonts}       
\usepackage{nicefrac}       
\usepackage{microtype}      

\usepackage{float}
\usepackage{graphicx}
\usepackage{subfig}

\usepackage{cite}

\title{Hierarchical BiGraph Neural Network as Recommendation Systems}

\author{
  \textbf{Dom Huh} \\ 
  $^{1}$Department of Electrical and Computer Engineering, George Mason University, Virginia, USA \\
  \texttt{dhuh4@gmu.edu}
}

\begin{document}

\maketitle

\begin{abstract}
Graph neural networks emerge as a promising modeling method for applications dealing with datasets that are best represented in the graph domain. In specific, developing recommendation systems often require addressing sparse structured data which often lacks the feature richness in either the user and/or item side and requires processing within the correct context for optimal performance. These datasets intuitively can be mapped to and represented as networks or graphs. In this paper, we propose the Hierarchical BiGraph Neural Network (HBGNN), a hierarchical approach of using GNNs as recommendation systems and structuring the user-item features using a bigraph framework. Our experimental results show competitive performance with current recommendation system methods and transferability.
\end{abstract}

Recommendation systems have largely been handled under two paradigms: collaborative filtering and content-based filtering. The former leverages correlation amongst a population of users and their observable recommendations to predict ratings \cite{Goldberg} whereas the latter utilizes information retrieval systems for analysis on the past users and items relationship to formulate the rating prediction as a supervised learning task \cite{Lang}. The unification of these two paradigm have previously been successfully explored in past works to find that they are inherently complementary to each other as they mitigate each other's limitations. For example, \cite{Balabanovic} implemented FAB, an hybrid recommendation system, which uses content-based filtering to develop user profiles based on their past ratings to recommend similar items, and collaborative filtering to broaden the scope of the user to similar users for unseen items. Whereas in pure systems, they are limited in the ability to do the other's task.

In terms of data, recommendation system tasks often require processing largely sparse contextual information on both the user and item ends. Intuitively, the information can be structured as some network of users and/or items to encapsulate the contextual relationships. The challenge of how to process user and item information has also raised some consideration: whether to compute using user and item independently \cite{Zhao, Sarwar} or to extract features from user and item jointly \cite{Basilico}. Past works have typically used some form of a bipartite graph \cite{Zhang, Berg}, but in this paper, we propose using a generalized bigraph framework, which adopt some formalism discussed in past works on pure and sharing bigraphs \cite{Milner, Sevegnani}. This framework allows us to handle the information locally and globally in a modular manner, a property explored in past works \cite{Wang}.

In this paper, we introduce the Hierarchical BiGraph Neural Network (HBGNN) as a recommendation system. HBGNN implicitly utilizes both collaborative filtering and content-based filtering, which is discussed further in Section.\ref{HBGNN}, and partitions processing using graph neural networks (GNNs) \cite{Scarselli, Bruna, Duvenaud, Dai} of user and item using a hierarchical graph framework. HBGNN is able to use node-level GNNs to create user and item profile within the subgraphs, and a graph-level GNN on the user and item profile to capture the contextual information within the root graph. Then, this contextual information can be processed using a traditional feed forward network to generate an rating. 

\section{Preliminaries} \label{prelim}
To test our model, we experimented on the MovieLens 100K dataset created by the GroupLens Research Project at the University of Minnesota \cite{Maxwell}, which is a common benchmark dataset for recommendation systems. It consists of 100,000 ratings on a scale of 5 from 943 users on 1682 movies. Along with unique identification numbers, for each user, the age, occupation, zip code and gender are provided, and for each movie, the movie title, genre, and other metadata are provided. For our experiments, we use the unique identification number, age, occupation, zip code and gender as user features, and the unique identification number and genre as movie features. For this dataset, there exists a standard 5-fold training and validation split which we used in order to accurately compare with current state of the art models. To test for transferability, we experimented on the MovieLens 1M dataset, which has 1 million ratings from 6000 users on 4000 movies and has the same user and item metadata. As there is no defined train-valid split for this dataset, we used a 80-20 train-valid split, with the latter timestamps of the ratings held out as validation.

\begin{figure}[H]
  \centering
  \begin{minipage}[b]{40mm}
    \includegraphics[width=40mm]{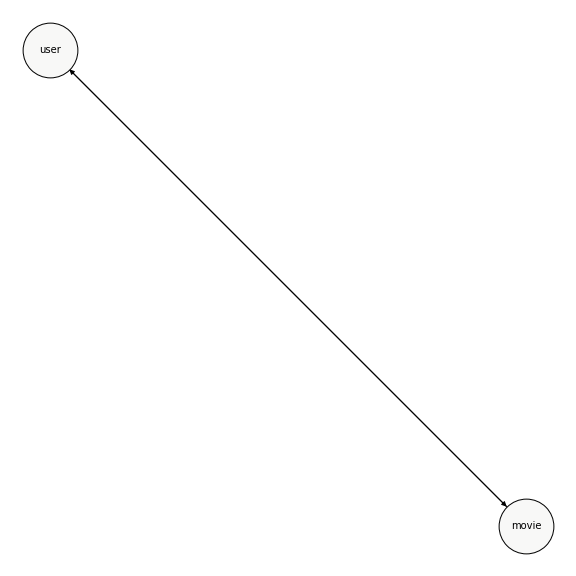}
    \caption{Rating Graph}
    \label{rg}
  \end{minipage} 
  \quad \quad \quad \quad \quad
  \begin{minipage}[b]{40mm}
    \includegraphics[width=40mm]{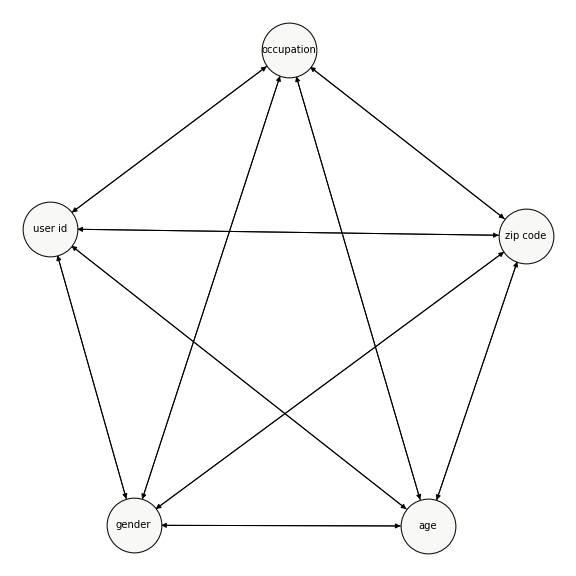}
    \caption{User Graph}
    \label{ug}
  \end{minipage}
  \begin{minipage}[b]{40mm}
    \includegraphics[width=40mm]{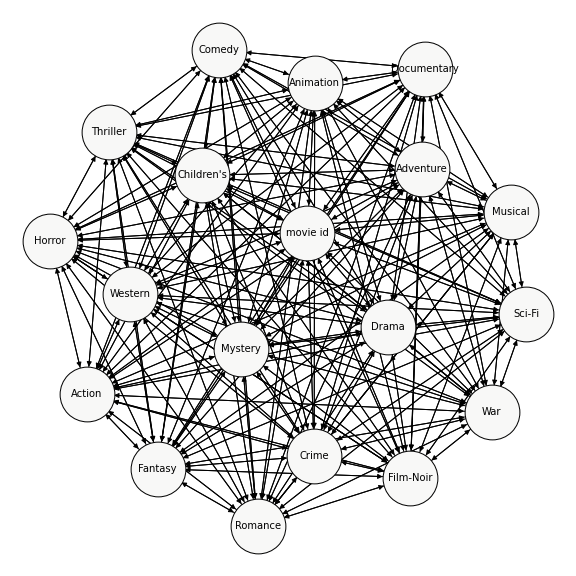}
    \caption{Movie Graph}
    \label{mg}
  \end{minipage}
  \caption{\textbf{MovieLens Graphs}: The three graphs above specify the graphs used for MovieLens 100K for $\alpha$-HBGNN. For $\beta$-HBGNN, the deviation would be the elimination of the identification number nodes on both user and movie level graphs, and introducing these features at the rating place graph in their respective nodes.}
  \label{movielengraph}
\end{figure}

With the HBGNN described in Section.\ref{HBGNN}, we have formulated a supervised learning task with a defined $x$ as the user-item bigraph and $y$ as the rating. Thus, to tune our model, we use root mean squared error, $RMSE$, as our cost function given our prediction, $\hat{y}$, and the target value, {y}.
\begin{equation}
    RMSE = \sqrt{\sum^{n}_{i}{\frac{(y-\hat{y})^{2}}{n}}}
\label{rmse}
\end{equation}
We use AMSGrad with Weight Decay \cite{Sashank, Krogh}, an adaptive gradient descent optimization based on exponential moving average updates paired with weight regularization, to optimize the cost function defined in Eq.\ref{rmse}.

\section{Hierarchical BiGraph Neural Network (HBGNN)}
\label{HBGNN}

\begin{figure}[H]
\centering{\includegraphics[width=0.7\textwidth]{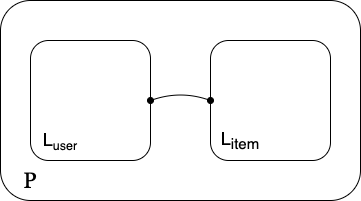}}
\caption{Generalized bigraph framework: Example with user and item in context of recommendation system with a single port. }
\label{graph}
\end{figure}

We start by formalizing the generalized bigraph framework used to structure the user-item features of the recommendation system using graphical form shown in Fig. \ref{graph}. The generalized bigraph framework introduced allows us to sufficiently express the locality and connectivity of the entities and its features. In the context of recommendation systems, there exists two entities: user and item. Thus, there are two nodes placed in the scope of the place graph, $P$, and is connected by a single port. This port provides the communication between the two entities needed to merge the two entities for the task at hand. For recommendation systems, the task is to predict the ratings given user and item. The bigraph needs to be concrete, which means all supports need to be defined. Thus, to obtain the state for the two entities in the place graph, we define two link graphs that contain a densely connected nodes with states allocated by features. The links amongst the peers are bidirectional, and communicate to embed the profile of the entity. 

\begin{figure}[H]
\centering{\includegraphics[width=0.7\textwidth]{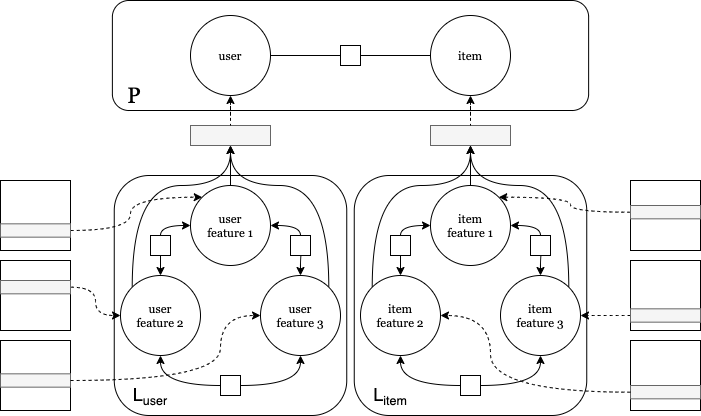}}
\caption{Bigraph Deconstruction: Example of a deconstructed bigraph to clearly illustrate the locality and connectivity of the generalized bigraph framework in context of recommendation systems and graph neural networks under the assumption of three features for both user and item entities. }
\label{detailedgraph}
\end{figure}

Now, we will canonically go over the steps of how the HBGNN propagates the user and item features to predict a rating using Fig. \ref{detailedgraph}. Since features of the user, $f_u$, and item, $f_i$, are sparse and usually discrete, we introduce learnable embeddings, $e_u$ and $e_i$, to be assigned to their respective node within the link graphs. 
\begin{equation}
    f_u \longrightarrow e_u, \quad f_i \longrightarrow e_i
\label{fe}
\end{equation}

Once embeddings are assigned to their respective node, both link graphs perform message passing. Each node processes incoming messages using an gated recurrent unit \cite{Li} with the input $x$ being the aggregated sum of all $n$ neighbor messages, $m_{i,\{1...n\}}$, for node $i$ denoted as $M_i$ and the hidden state $h$ being the current state denoted as $h_{i,t-1}$, at each updated node feature, $h_{i,t}$. It is important to note that the message sent by neighbors are the current state of the neighbors, and no independent transforms should be applied individually to each message rather on the aggregated sum as to be tolerant to isomorphism. The message passing process can done concurrently through all nodes in both link graphs, thus being very computational efficient.

\begin{equation}
    h_{i,t+1} = \textit{GRU}(x = M_{i,t}, h = h_{i,t}) \\
    \textit{where} \quad M_{i,t} = \sum^n_k m_{i,t,k} \quad \textit{for some node i}
\label{mp}
\end{equation}

Upon $t$ messages passing rounds, all node states in the link graphs are passed through an neural network for encapsulation and then assigned to the respective nodes, $N_u$ and $N_i$, in the place graph. With our design choice of making the link graph fully connected, $t$ can equal 1 and still allow propagation through all combinations of nodes states. We will experiment with this claim in Section \ref{result}.
\begin{equation}
    N_u = \theta_u \cdot H_u, \quad N_i = \theta_i \cdot H_i
\label{ff}
\end{equation}

Since all supports are defined upon assignment, we can communicate via port for the recommendation system task. We can perform message passing between the two nodes at the place graph similar to how it was done in the link graphs. Again, each entity node contains a gated recurrent unit and messages are aggregated in the same manner. We can additionally choose whether the identification number of the user and item is introduced in the link graph or the place graph. Models that input the identification number at the link graph will be prefixed with $\alpha$ whereas models that input the identification number at the place graph will be prefixed with $\beta$. Furthermore, once again, since the place graph is fully connected, complete propagation between the two peers is done with a single iteration of message passing.

The priors to determine the sparsity in the link graphs $G_L = \{V_L,E_L\}$ are not trivial. In graph structures, its sparsity can be controlled by either edge removal or edge reweighing. We argue edge reweighing is a more generalized approach of edge removal, since given $e \in E_L$ that connects node $v_1$ to node $v_2$, the message passed through $e$ can be scaled by 0 to represent edge removal. To formalize edge reweighing, we denote $A$ as the set of edge weights, and for all edges in $G_L$, there is an associated element in $A$. Thus, $| A | = | E_L |$. Now, message passing with edge reweighing can be a weighted sum of all neighboring messages.

\begin{equation}
    M_i = \sum^n_k a_{i,k} * m_{i,k} \quad \textit{for some node i}
\label{edgerw}
\end{equation}

To compute each $a \in A$, we can leverage past works \cite{vel, Vaswani, Liao} on the attention mechanism and graph neural networks. Simply, we can assign the hidden state $h_i$ as the key, and each message $m_{i,\{1...n\}}$ as the query and value for node $i$. 

\begin{figure}[H]
\centering{\includegraphics[width=0.6\textwidth]{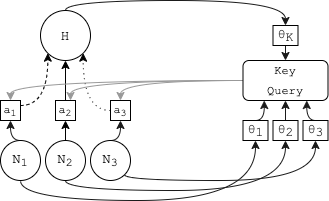}}
\caption{Edge reweighing with attention mechanism with a node $H$ with three neighbor nodes. }
\label{attentionfig}
\end{figure}

The key and query are transformed using a linear transform $\theta_k$ and $\theta_q$.

\begin{equation}
    \textit{score}_i = (\theta_k \cdot h_i)^T \cdot (\theta_q \cdot m_{i,\{1...n\}}) \\
    a_{i,j} = \frac{\textit{exp}(\textit{score}_{i,j})}{\sum^n_k \textit{exp}(\textit{score}_{i,k})} \quad \textit{where} \quad j \in [1,n]
\label{attention}
\end{equation}

Thus, we introduce a variant of the HBGNN that utilizes the attention mechanism for edge reweighing, as to overcome the unknown prior for feature dependency and sparsity in both user and item link graphs. We denote this model as Attention HBGNN (AHBGNN).

To obtain the rating, we can proceed in numerous ways by utilizing some combination of the nodes of the link graph and/or place graph, but we will only illustrate on using solely the place graph in this paper. Once message passing is complete, we can pass the states of the two entity nodes of the place graph into an neural network to predict the rating. To optimize the HBGNN, we can treat the recommendation system problem as a supervised learning task. 

\begin{table}[]
\begin{center}
\begin{tabular}{c|c|c|c|c}
\hline
Methods & \multicolumn{4}{c}{RMSE}                                       \\
& \multicolumn{2}{c|}{MovieLens 100K} & \multicolumn{2}{c}{MovieLens 1M} \\
& Train & Test & Train & Test \\ \hline
GCMC & - & 0.910 & - & 0.832\\
IGCMC & - & 1.142 & - & 1.259\\
IGMC & - & 0.905 & - & 0.857\\
PinSage & - & 0.951 & - & 0.906 \\
FEAE & - & 0.920 & - & 0.860 \\
SSM & - & 0.910 & - & 0.863 \\
MLP & 1.141 & 1.178 & 1.123 & 1.149 \\
$\alpha$-HBGNN  & 0.002 & 0.927 & 0.002 & 0.877 \\
$\alpha$-HBGNN*  & 0.002 & 0.914 & 0.003 & 0.863 \\
$\beta$-HBGNN  & 0.729 & 0.930 & 0.581 & 0.898 \\
$\beta$-HBGNN*  & 0.704 & 0.912 & 0.579 & 0.879 \\
$\alpha$-AHBGNN  & 0.001 & 0.910 & 0.002 & 0.852 \\
$\beta$-AHBGNN  & 0.708 & 0.931 & 0.548 & 0.870 \\
\hline
\end{tabular}
\caption{RMSE train-test results on MovieLens 100K and 1M. The suffix * denotes transfer learning.}\label{tab:result}
\end{center}
\end{table}

We argue that HBGNN is implicitly utilizing both collaborative filtering and content-based filtering under this generalized bigraph paradigm. HBGNN embeds the user and item profiles into disjoint graphs, encoding the same information used to describe by all the users and items in the user and item population. The learned embedding is generalized and optimized in a manner that take into account other users and item, and this aspect acts as collaborative filtering since it can generate correlations amongst the population. By using a supervised learning structure, HBGNN utilizes on past user and item relationship. Thus, this aspect acts as content-based  filtering since it can generate correlations based on user-item history. 

\section{Results} \label{result}
We conducted experiments described in Section \ref{prelim}, and compare our models, $\alpha$-HBGNN, $\beta$-HBGNN, and AHBGNN, to the current modeling methods that were tested on both datasets on the RMSE metric, such as Graph Convolution Matrix Completion (GCMC) \cite{Berg}, Inductive Graph Convolution Matrix Completion (IGCMC) \cite{Zhang}, Inductive Graph-based Matrix Completion (IGMC) \cite{Zhang}, PinSage \cite{Ying}, Self-Supervised Exchangeable Model (SSEM) \cite{Hartford}, and Factorized Exchangable Autoencoder (FEAE) \cite{Hartford}. Also, we tested a traditional multi-layer perceptron model (MLP) as well on the datasets with the same features as the HBGNNs and their own set of learnable embedding.

Both $\alpha$-HBGNN and $\beta$-HBGNN uses 512 parameters for each state in the user and movie link graphs, and 2048 parameters for each state in the rating place graph. The size of the embedding matrix depends on the range of the user and movie feature, but we set the user age range to be fixed at 100 so that we can leverage transfer learning from MovieLens100K and MovieLens1M. The encoding neural network that assigns the state of the nodes in the place graph is a single layer model with 4096 linear perceptrons. The processing network used to predict the rating is a 5-layer multi-layer perceptron model with leaky rectifier transforms between the layers. All linear layers are initialized with Kaiming initialization using an uniform distribution.

The results shown in Table \ref{tab:result} conclude our HBGNNs to performs competitively with the current modeling methods on both dataset. The $\alpha$-HBGNN and $\beta$-HBGNN was trained for 75 epochs on the MovieLens100K dataset, and trained for 30 epochs for MovieLens1M, and the final RMSE loss is shown in Table \ref{tab:result}. The attention variants followed respective training procedures.

Additionally, we fine-tuned the pretrained $\alpha$-HBGNN and $\beta$-HBGNN that performed the best amongst the five folds from the MovieLens100K dataset on the MovieLens1M dataset, and the pretrained models outperforms the models without pre-training within one epoch. We also attempted fine=tuning the pretrained $\alpha$-HBGNN and $\beta$-HBGNN from the MovieLens1M dataset on the MovieLens100K dataset, and found similar results. Both results can be seen in Table \ref{tab:result} and each pretrained model was only fine-tuned for 5 epoch for both datasets, as it converged very quickly and to a better local optima. To do this, we had to reformat and retrain the unique identification number embeddings for user and movie as they differed between the two datasets, as well the zip code for the user profile. So, when we fine-tune the models, HBGNN only needs to learn the embeddings for the identification numbers and the zip code, and adjust the ports and links minimally. From our experimental results in Table \ref{tab:result}, we can conclude the transferability of this architecture to be very effective, only required to relearn the embeddings for the identification number and the zip code for the user.

\begin{figure}
\centerline{\includegraphics[width=\columnwidth]{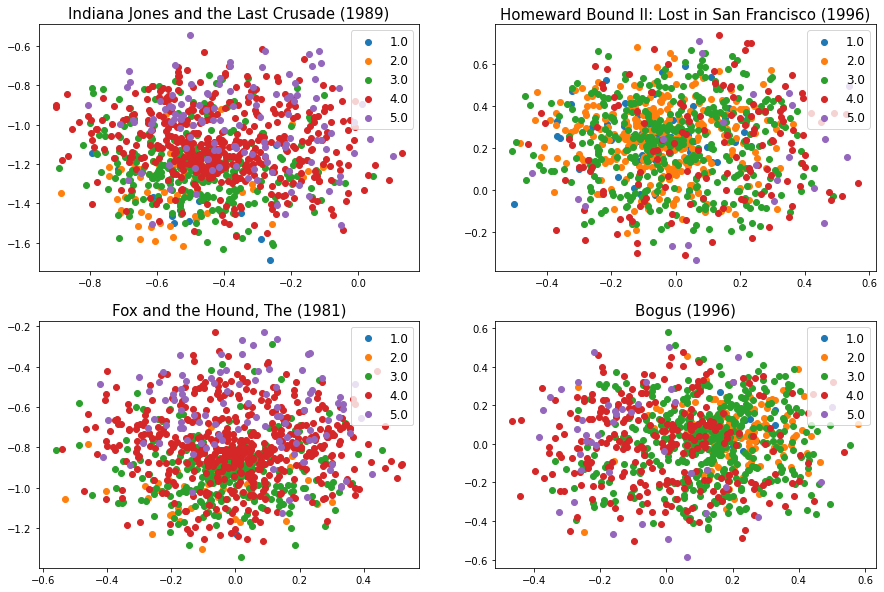}}
\caption{TSNE visualization of user profiles using Chebyshev distance distinguished by their respective ratings.} \label{fig:tsne}
\end{figure}

The $\alpha$-HBGNN achieved faster convergence than $\beta$-HBGNN, however once converged, both models achieved similar performance on validation set, with $\alpha$-HBGNN performing slightly better. $\alpha$-HBGNN appears to overfit to the training set much more than $\beta$-HBGNN in both MovieLens datasets. Both methods achieve competitive performance amongst current methods, and additional regularization should be applied in future works for better generalization and performance on the validation set.

In Fig. \ref{fig:tsne}, we visualize each user profiles created by the user link graph and their respective ratings generated by the place graphs using t-distributed Stochastic Neighbor Embedding (TSNE) using $\alpha$-HGNN with 1 message passing round for both place and link graphs. The details of each movie is listed in Table \ref{tab:movie}.

\begin{table}
\begin{tabular}{l|l|c|c|c|c|c}
\hline
\multicolumn{1}{c|}{Movie Titles} & \multicolumn{1}{c|}{Genre} & \multicolumn{5}{c}{Number of Ratings} \\
\multicolumn{1}{l|}{} & \multicolumn{1}{l|}{} & \multicolumn{1}{l|}{1} & \multicolumn{1}{l|}{2} & \multicolumn{1}{l|}{3} & \multicolumn{1}{l|}{4} & \multicolumn{1}{l}{5} \\ \hline
Star Trek: First Contact (1996) & Action, Adventure, Sci-Fi & 5 & 25 & 87 & 123 & 53 \\ \hline
Homeward Bound II... (1996) & Adventure, Children's & 6 & 8 & 6 & 8 & 2 \\ \hline
Cats Don't Dance (1997) & Animation, Children's, Musical & 5 & 6 & 7 & 1 & 5 \\ \hline
Love Bug, The (1969) & Children's, Comedy & 4 & 8 & 24 & 4 & 1 \\ \hline
\end{tabular}
\caption{Movies used for TSNE visualization, showing genre and distributions of provided ratings within MovieLens100K dataset.} \label{tab:movie}
\end{table}

\section{Conclusion}
In this paper, we have proposed Hierarchical BiGraph Neural Net (HBGNN), a hierarchical approach of using graph neural networks as recommendation systems and structuring the user-item features using a bigraph framework. We show that HBGNN is competitive compared to current recommendation systems and is transferrable to higher scaled tasks without much retraining. In future works, we hope to explore different methods of understanding the embeddings in the link and place graphs of the HBGNN (ie. understand the individual and joint profiles), and test different design choices, (ie. methods of message passing, connectivity, different graph neural network architectures).

\bibliography{reference}{}
\bibliographystyle{plain}

\end{document}